\documentclass[sigconf,nonacm]{acmart}
\AtBeginDocument{%
  }

\setcopyright{none}

\renewcommand\footnotetextcopyrightpermission[1]{} 

\newcommand{\dataset}{PA-EgoQA}

\newcommand{\filter}{\textsc{Filter-GRPO}}

\usepackage{booktabs}
\usepackage{multirow}
\usepackage{xcolor}
\usepackage{colortbl}
\usepackage{tabularx}
\usepackage{makecell}
\usepackage{fontawesome}          
\newcommand{\mch}[1]{\multicolumn{1}{c}{\textbf{#1}}}
\newcommand{\yes}{\ensuremath{\checkmark}}
\newcommand{\no}{\textcolor{black!45}{\ensuremath{\times}}}

\newcolumntype{Y}{>{\centering\arraybackslash}X}

\definecolor{gradBase}{HTML}{2A9D8F}
\newcommand{\gOne}{\cellcolor{gradBase!8}}
\newcommand{\gTwo}{\cellcolor{gradBase!15}}
\newcommand{\gThree}{\cellcolor{gradBase!22}}
\newcommand{\gFour}{\cellcolor{gradBase!30}}
\newcommand{\gFive}{\cellcolor{gradBase!40}}
\newcommand{\corr}{\raisebox{0.5ex}{\scalebox{0.62}{\faEnvelope}}}
\begin{document}

\title{PAVE: Preference-Aligned Visual Egocentric Learning for the Blind and Low-Vision Users}
\author{\texorpdfstring{%
  Nan Li\textsuperscript{$\star$,1},\
  Shaofeng Liang\textsuperscript{$\star$,$\clubsuit$,1,2},\
  Shiyun Zhao\textsuperscript{$\star$,1},\
  Keni Shi\textsuperscript{1},\
  Zhencheng Huang\textsuperscript{1},\\
  Bowen Tian\textsuperscript{1,2},\
  Jiemin Wu\textsuperscript{1},\
  Wenshuo Chen\textsuperscript{1},\
  Kaishen Yuan\textsuperscript{1},\
  Runwei Guan\textsuperscript{1},\
  Yutao Yue\textsuperscript{\corr,1,3}%
}{}}
\affiliation{%
  \institution{\textsuperscript{1}The Hong Kong University of Science and Technology (Guangzhou)\\
  \textsuperscript{2}EigenAct Technology Co., Ltd. \
               \textsuperscript{3}Institute of Deep Perception Technology (JITRI)}\      
    \textsuperscript{$\star$}Equal contribution,\quad
               \textsuperscript{$\clubsuit$}Project Lead,\quad
               \corr\,Corresponding Author   
                \city{ }
  \country{ }
}

\renewcommand{\shortauthors}{Nan Li et al.}

\begin{abstract}
Egocentric video question answering enables blind and low-vision (BLV) users to query their immediate surroundings through natural language. However, existing benchmarks and models primarily optimize for semantic agreement with reference answers, reducing visual assistance to a purely factual matching task. In real-world scenarios, a factually correct response remains practically unhelpful if it lacks the spatial orientation, actionable detail, or conciseness required by the immediate task of the user. Drawing upon real BLV egocentric videos and authentic user queries from EgoBlind, we construct the PA-EgoQA, a preference-aligned dataset that reconstructs reference answers according to empirical BLV communication criteria. Based on this dataset, we introduce Preference-Aligned Visual Egocentric Learning (PAVE), a unified learning and inference framework that couples structured answer-policy fine-tuning with deterministic question-conditioned routing to adapt response formulation directly to the surface form of the query at inference time. We further align the model via Filter-GRPO, which pairs a hierarchical correctness-quality reward with contrast-based candidate-group filtering to extract learning signals from both factual accuracy contrasts and fine-grained quality variations among correct answers. Extensive experiments across three vision-language backbones demonstrate consistent improvements in factual accuracy alongside substantial gains in assistive response quality. These results establish a concrete paradigm for translating human-centered response requirements into egocentric supervision, generation policies, and post-training objectives. Code will be provided at \href{https://github.com/shawnliang420/PAVE}{https://github.com/shawnliang420/PAVE}. \par\vspace*{-1\baselineskip}
\end{abstract}



\keywords{Blind and Low-Vision Assistance, Egocentric VQA, Web Accessibility, Preference Alignment, Vision-Language Models}

\begin{teaserfigure}
\includegraphics[width=\textwidth]{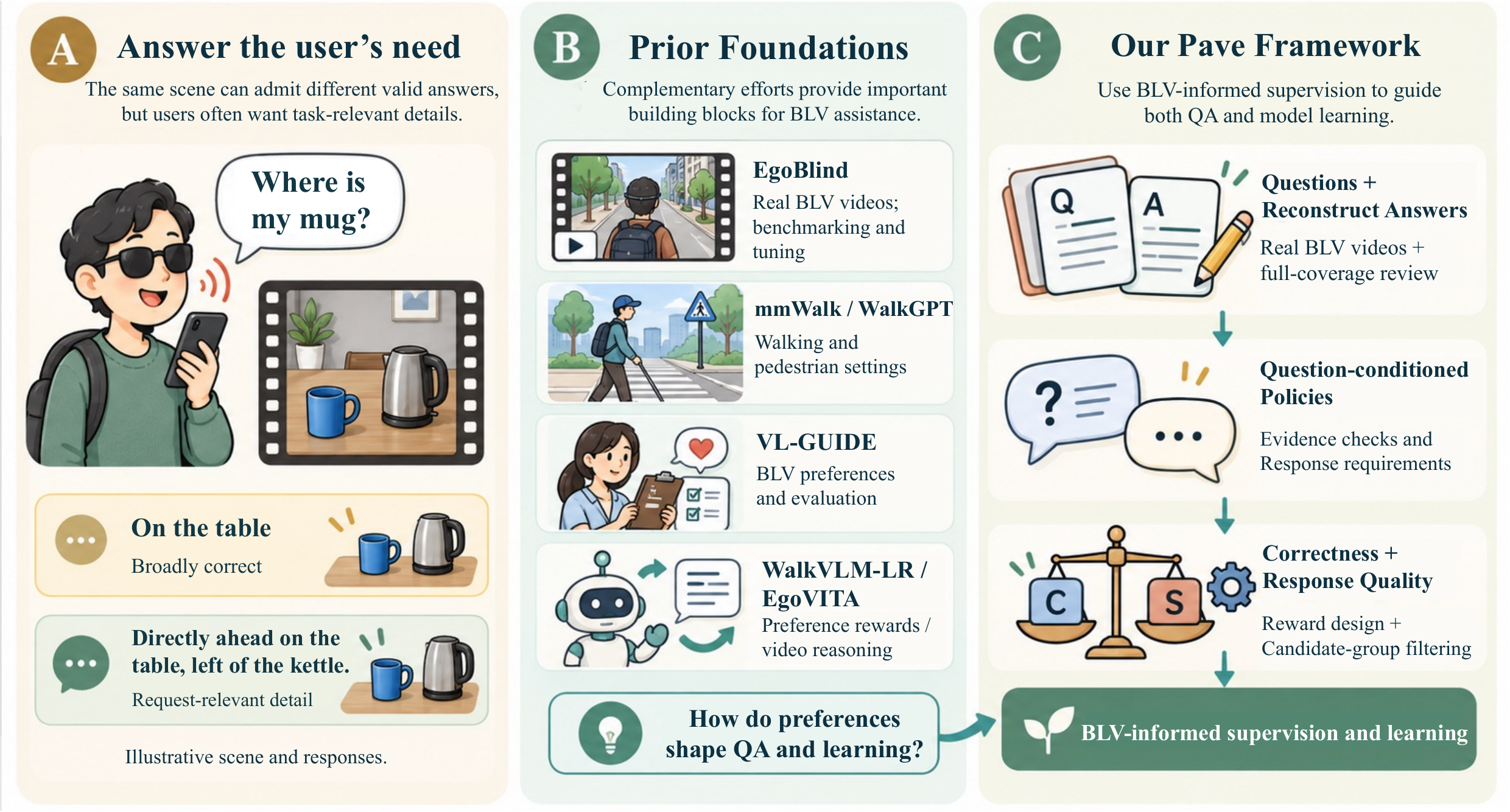}
\caption{Conceptual overview of preference-aligned egocentric visual assistance. (A) Blind and low-vision (BLV) users require task-relevant spatial details rather than broadly correct generic answers. (B) Prior studies establish essential foundations across egocentric benchmarks, navigation, preference criteria, and video reasoning. (C) Our framework integrates BLV-informed supervision to jointly guide QA reconstruction, question-conditioned policies, and reinforcement learning.}
\vspace{1em}
\label{fig:motivation}
\end{teaserfigure}

\maketitle

\section{Introduction}
\label{sec:intro}

Multimodal Web accessibility services increasingly rely on natural-language responses as an interface between users and visual information. For blind and low-vision (BLV) users~\citep{zheng2026insight,zheng2024materobot}, open-ended egocentric video question answering (VideoQA) offers a vital mechanism to request situated assistance. Typical requests range from reading a display or locating an object to operating a control or interpreting a dynamic scene~\citep{zhang2021trans4trans,hwang2024safe}. In such interfaces, generated responses serve as an accessible representation of visual evidence. Our key insight is that factual correctness and task-specific assistive utility are fundamentally distinct: \textbf{an answer can be visually accurate yet practically unhelpful if it omits the precise spatial, procedural, or factual granularity demanded by the immediate task of the user.} Reconciling this discrepancy between semantic accuracy and situated communication quality constitutes a central challenge for user-centered multimodal Web accessibility~\citep{hu2026position,zheng2026insight}.

Although recent studies provide essential building blocks for this domain~\citep{hwang2026navable,hwang2026guidetwsi,hwang2024synthetic}, existing resources and human preference findings remain disconnected. On the benchmark side, EgoBlind~\citep{xiao2026egoblind} curates first-person videos from the daily lives of BLV individuals alongside authentic user-verified questions. This resource enables the study of situated visual assistance well beyond conventional scene captioning. Nevertheless, its reference annotations and evaluation protocols focus primarily on semantic agreement rather than assistive communication quality. Conversely, human-centered studies on BLV assistance establish the importance of structured organization, explicit user-relative spatial cues, and a careful balance between informational sufficiency and conciseness~\citep{kim2025large,zhao2025less}. However, these studies stop short of translating empirical guidelines into actionable learning signals. Consequently, operationalizing human preference principles across training supervision, inference routing, and post-training optimization for open-ended egocentric VideoQA remains an open problem.

Based on this observation, we formulate assistive response generation as a preference-aligned decision process and introduce \textbf{PAVE} (\textbf{P}reference-\textbf{A}ligned \textbf{V}isual \textbf{E}gocentric Learning), a unified framework for preference-informed learning and single-pass routed inference. Diverse assistive tasks impose fundamentally different structural demands. For instance, a text-reading query calls for an exact factual readout, whereas an object-localization or procedural query requires ego-relative orientation or ordered action steps. To address these needs, PAVE unifies reference supervision, query-conditioned policy selection, and post-training optimization under a shared set of BLV communication requirements. This end-to-end alignment enables the system to generate task-appropriate final answers directly from visual evidence.

Specifically, PAVE operates across three connected stages. First, we construct \dataset{}, a preference-aligned egocentric VideoQA dataset, by reconstructing the reference answers of EgoBlind around six BLV-informed criteria: Direction Accuracy, Depth Accuracy, Safety and Actionability, No Hallucination, Sufficiency, and Conciseness. Four reviewers conduct a full-coverage, video-grounded verification of the reconstructed QA pairs and original question intent. Second, we couple structured answer-policy fine-tuning with a deterministic question-form router. This design adapts response formulation directly to the surface form of the query at inference time without introducing an auxiliary routing network. Third, we align the model via \filter{}, which pairs a hierarchical correctness-first reward with contrast-based candidate-group filtering. Groups containing both correct and incorrect answers support factual discrimination, whereas groups of exclusively correct answers with varying quality scores provide a clean learning signal for refining assistive expression. At deployment, the resulting system operates via single-pass greedy decoding, requiring neither reward-model calls nor multi-candidate voting.

Extensive experiments on \dataset{} demonstrate that PAVE consistently outperforms matched EgoBlind baselines across all three primary backbones, improving factual accuracy by 4.12\%, 4.30\%, and 0.83\%, respectively. Furthermore, PAVE achieves substantial gains in assistive quality while reducing response length by up to 50.8\%. By translating human-centered communication preferences into concise, spatially grounded, and actionable guidance under low-latency single-pass inference, PAVE lowers the cognitive load of real-world visual assistance. Ultimately, our framework empowers BLV individuals to navigate and interact with their daily environments with greater independence and safety.

Our primary contributions are summarized as follows:
\vspace{-1em}
\begin{itemize}
  \item \textbf{Preference-aligned egocentric QA dataset (\dataset{}).} We construct \dataset{}, a preference aligned dataset grounded in authentic BLV egocentric videos and user queries from EgoBlind. It reconstructs reference answers around six empirical BLV communication criteria, verified via a full-coverage four-reviewer video audit.
  \item \textbf{Unified learning and single-pass routed framework (PAVE).} We design PAVE to integrate structured answer-policy supervision and deterministic question-form routing with \filter{}. This post-training method pairs a hierarchical correctness-quality reward with contrast-based candidate-group filtering, while preserving single-pass greedy decoding with zero routing overhead.
  \item \textbf{Comprehensive validation across three VLM backbones.} Extensive evaluations on InternVL2.5-8B, LLaVA-OneVision-7B, and Qwen2.5-VL-7B demonstrate that PAVE consistently elevates factual accuracy and BLV assistive quality over matched baselines, while compressing response length by up to 50.8\% without additional inference latency.
\end{itemize}

\section{Related Work}
\label{sec:related_work}

\subsection{Visual Assistance for Blind and Low-Vision Users}

Visual question answering (VQA) empowers BLV individuals to acquire visual information about their surroundings through natural-language queries~\citep{gao2026visassist,kim2026guidedog}. Pioneering this direction, VizWiz anchored VQA in authentic accessibility scenarios by collecting photographs and spoken questions directly from blind users~\citep{gurari2018vizwiz}. Subsequent research expanded visual assistance into dynamic, temporal, and task-oriented domains. For instance, AssistQ explores procedural task guidance using egocentric instructional videos, whereas EgoTextVQA addresses multi-frame scene-text comprehension from a first-person perspective~\citep{WongCWLMGS2,zhou2025egotextvqa}. More recently, EgoBlind curated diverse daily assistance queries from first-person videos captured by BLV users~\citep{xiao2026egoblind}, and WalkVLM investigated continuous navigational support for blind pedestrians~\citep{yuan2025walkvlm}. 
Beyond wearable interfaces, embodied guide-dog robotics has also advanced human-centered design~\citep{hwang2024lessons,yu2026human}, visual navigation~\citep{hwang2026guidenav,liu2026not}, task assistance~\citep{liu2026robofind,yin2026guidefetch}, and VLM-based guidance~\citep{yun2026sa}. Despite these advances, existing datasets and models primarily optimize for factual agreement rather than the heterogeneous communication demands of BLV users. To address this limitation, we construct \dataset{} by reconstructing egocentric reference annotations around six empirical BLV criteria, embedding ego-relative spatial orientation, safety awareness, and conciseness directly into training supervision.

\subsection{Structured Reasoning and Reinforcement Learning for Multimodal Assistance}

Chain-of-thought (CoT) prompting enhances multi-step reasoning in large language models~\citep{wei2022chain}, and Multimodal-CoT extends this paradigm to vision-language tasks via visually grounded rationales~\citep{zhang2024multimodal}. Within accessibility, WalkVLM structures visual perception and navigational decisions through a hierarchical CoT pipeline~\citep{yuan2025walkvlm}. Nevertheless, uniform step-by-step reasoning is not universally advantageous. Prior studies demonstrate that while explicit CoT benefits complex reasoning, it can degrade direct perception tasks such as visual grounding and object counting~\citep{jin2026look}. This trade-off has motivated adaptive strategies such as AdaCoT, which selectively triggers reasoning based on query characteristics~\citep{lou2025adacot}.
In parallel, reinforcement learning, particularly Group Relative Policy Optimization (GRPO)~\citep{shao2024deepseekmath}, has emerged as an effective mechanism for aligning multimodal models through task-specific rewards and online rollout filtering~\citep{shen2025vlm,meng2025mm,zhao2026laf,choudhary2026small}. Building upon these insights, PAVE couples structured answer-policy supervision with deterministic question-conditioned routing, and introduces \filter{} to jointly learn from factual accuracy contrasts and fine-grained assistive quality variations.

\subsection{BLV-Centered Response Evaluation}

Standard VQA benchmarks evaluate outputs based on lexical or semantic overlap with references, which fails to capture whether a response delivers actionable and concise guidance for a BLV user~\citep{jiang2025bench,liang2026focus}. Although EgoBlind evaluates multimodal models across first-person queries using semantic matching, it does not decouple core factual correctness from assistive communication quality~\citep{xiao2026egoblind}. Recent human-centered research examines BLV user preferences and introduces accessibility-oriented evaluation criteria along dimensions such as sufficiency, actionability, and conciseness~\citep{kim2025large}. Similarly, WalkVLM-LR integrates preference rewards to suppress verbose descriptions during blind walking assistance~\citep{li2026less}. Advancing beyond isolated evaluation or domain-specific verbosity suppression, our work explicitly decouples core semantic correctness from holistic BLV assistive quality across reference reconstruction, question-conditioned policy routing, and hierarchical post-training optimization.

\section{\dataset{}: Preference-Informed Reference-Answer Reconstruction}
\label{sec:reconstruction}

The practical utility of an egocentric visual assistant is fundamentally bounded by the information that its training references teach it to convey. For BLV users, two candidate responses may agree on the central visual fact while differing substantially in the information required to complete a task. As illustrated in Figure~\ref{fig:motivation}, confirming the presence of a mug suffices for an existence query, whereas retrieving that mug requires an explicit ego-relative spatial relation. To embed such task-critical requirements into training supervision and evaluation standards, we introduce \dataset{}, a preference-informed reconstruction of the QA annotations in EgoBlind~\citep{xiao2026egoblind} that elevates each question--reference pair into a high-utility assistive learning target.

\begin{figure}[t]
\centering
\includegraphics[width=\linewidth,trim=0 0 0 10bp,clip]{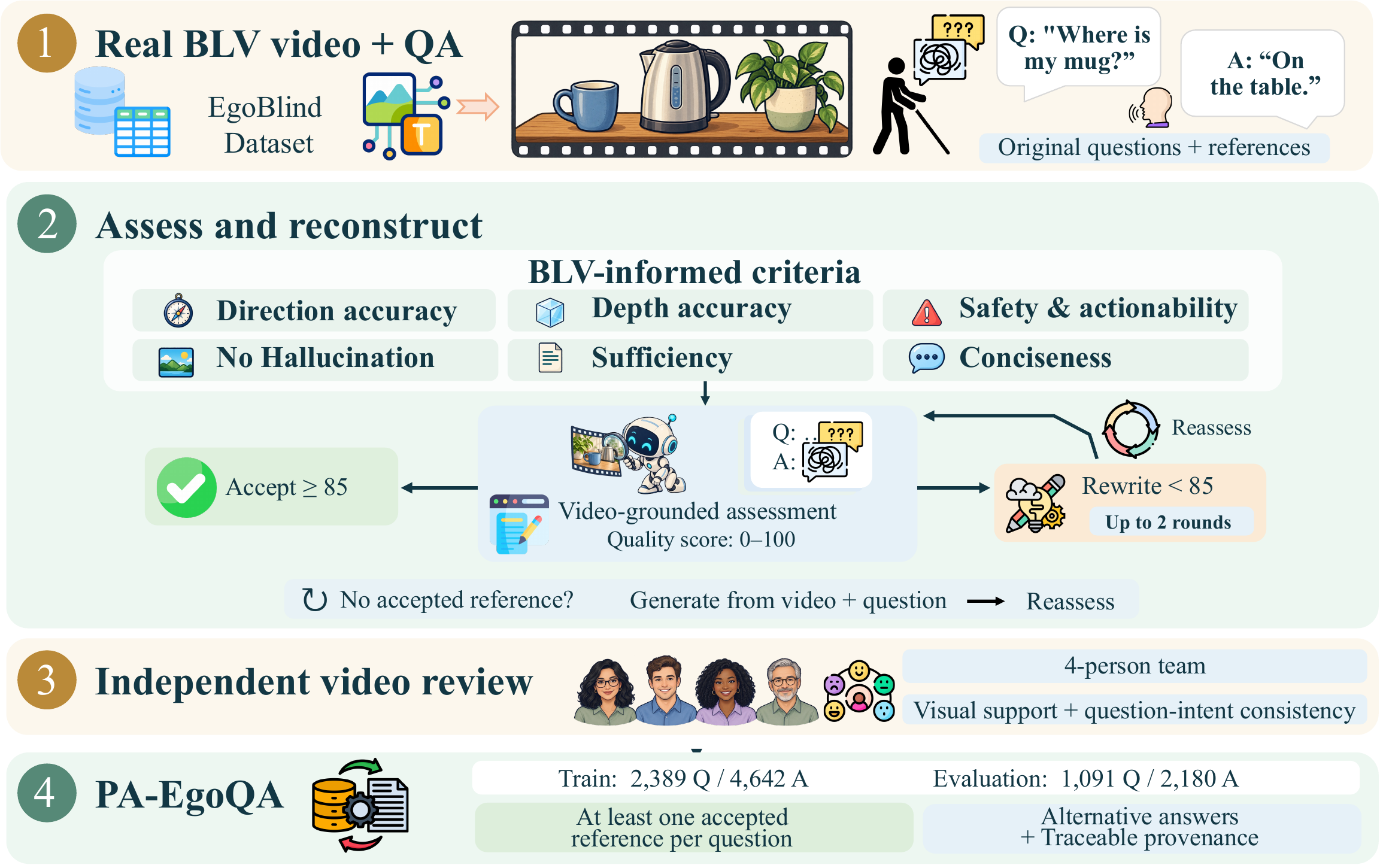}
\caption{Overview of the \dataset{} construction pipeline.}
\Description{A four-stage vertical flow chart from original EgoBlind video and QA data, through BLV-informed video-grounded assessment and reconstruction, four-reviewer independent video review, to the final dataset.}
\label{fig:reference_pipeline}
\end{figure}

\subsection{BLV-Oriented Reconstruction Criteria}
\label{sec:reconstruction_criteria}

Drawing upon empirical BLV preference studies~\citep{kim2025large}, we establish six complementary criteria for assistive response construction: \textbf{direction accuracy}, \textbf{depth accuracy}, \textbf{safety and actionability}, \textbf{no hallucination}, \textbf{sufficiency}, and \textbf{conciseness}. The salience of each criterion is conditioned on the query: information reading demands exact textual transcription, whereas object localization or appliance operation requires ego-relative directional cues or sequential steps. Across all categories, our overarching principle is \textbf{minimum sufficient assistance}: an answer must concisely communicate all factual and spatial details necessary to fulfill the stated task, without unsupported claims.

In egocentric assistance, the final frame of the video segment anchors the current scene state and the immediate perspective of the user, while preceding frames supply essential temporal evidence when targets appear clearer earlier or when motion trajectories and state transitions must be interpreted. Every reconstructed reference in \dataset{} remains spatially compatible with the terminal viewpoint, expresses explicit uncertainty when visual evidence is inconclusive, and treats multiple valid references for a single question as distinct \textbf{alternatives} rather than concatenating details, thereby preventing overly verbose generation habits.

\subsection{Video-Grounded Assessment and Reconstruction}
\label{sec:reference_revision}

Figure~\ref{fig:reference_pipeline} outlines the multi-stage reconstruction and human auditing pipeline used to construct \dataset{}. Each original reference is independently assessed alongside its question, task category, and temporally ordered sampled frames by a video-aware evaluator (gpt-4o-mini), which produces a holistic assistive quality score $J_{\mathrm{vid}}\in[0,100]$ and a structured rationale guided by the six BLV criteria. Original references that meet the strict threshold $\tau=85$ are directly retained. Lower-scoring references undergo up to two rounds of video-grounded revision and reassessment using a separate generator (gemini-3.6-flash) that has no access to the scores or rationales of the evaluator. If a question still lacks an accepted reference, a fallback candidate is generated directly from the video and question and is subsequently assessed under the same threshold.

Let $V_q$ denote the first-person video segment available when a BLV user poses a question $q$. The model generates an answer $a_q$ that satisfies the request under the current viewpoint. Let $\mathcal A_q^{\mathrm{cand}}$ denote the candidate pool for question $q$. The accepted reference set $\mathcal A_q^*$ and the question retention rule $\operatorname{Keep}(q)$ are defined as follows:
\begin{align}
\mathcal A_q^*&=\operatorname{Dedup}\!\left(\{a\in\mathcal A_q^{\mathrm{cand}}:
 J_{\mathrm{vid}}(V_q,q,a)\ge\tau\}\right),\label{eq:accept_ref}\\
\operatorname{Keep}(q)&=\mathbb I\big[|\mathcal A_q^*|\ge1\big].\label{eq:keep_q}
\end{align}
The deduplication operation normalizes letter case and whitespace while preserving the priority order of the candidates. Under the acceptance rule in Eq.~(\ref{eq:keep_q}), each retained question in \dataset{} has 1 to 4 accepted references (2 on average). Furthermore, a four-person review team completed a full-coverage, independent video-based verification of the reconstructed QA pairs, validating both visual support and the preservation of the original question intent. Source question identifiers, candidate references, assessment outcomes, revision histories, reviewer provenance, and final acceptance decisions remain fully traceable throughout reconstruction.
\begin{table}[t]
\centering
\caption{Comparison of egocentric and BLV-oriented QA datasets.
\textbf{\#V}: unique clips; \textbf{VLen}: mean clip duration (s); \textbf{\#Q}: questions;
\textbf{BV}: BLV-captured visuals; \textbf{BQ}: BLV-verified queries;
\textbf{RT}: real-time evidence; \textbf{MA}: multiple references;
\textbf{BPA}: BLV-preference alignment; \textbf{OE}/\textbf{MC}: open-ended/multiple-choice.}
\label{tab:dataset_comparison}

\fontsize{8}{9.6}\selectfont
\setlength{\tabcolsep}{2pt}
\renewcommand{\arraystretch}{1.15}
\begin{tabular}{l r r r c @{\hspace{2pt}} c c c c c}
\toprule
& \multicolumn{3}{c}{\itshape Scale} & & \multicolumn{5}{c}{\itshape BLV-oriented} \\
\cmidrule(lr){2-4}\cmidrule(lr){6-10}
\textbf{Dataset} & \mch{\#V} & \mch{VLen} & \mch{\#Q} & \mch{Type}
& \mch{BV} & \mch{BQ} & \mch{RT} & \mch{MA} & \mch{BPA} \\
\midrule
\multicolumn{10}{l}{\itshape General-purpose} \\
EgoTaskQA~\citep{jia2022egotaskqa}
& 2,000 & 25.0 & 40,000 & OE & \no & \no & \no & \no & \no \\
EgoSchema~\citep{mangalam2023egoschema}
& 5,031 & 180.0 & 5,031 & MC & \no & \no & \no & \no & \no \\
\midrule
\multicolumn{10}{l}{\itshape Assistive tasks} \\
AssistQ~\citep{WongCWLMGS2}
& 100 & 115.0 & 531 & MC & \no & \no & \no & \no & \no \\
EgoTextVQA~\citep{zhou2025egotextvqa}
& 1,507 & 101.7 & 7,064 & OE & \no & \no & \yes & \no & \no \\
\midrule
\multicolumn{10}{l}{\itshape Blind and low-vision assistance} \\
VizWiz~\citep{gurari2018vizwiz}
& \multicolumn{1}{c}{--} & \multicolumn{1}{c}{--} & 31,173 & OE
& \yes & \yes & -- & \yes & \no \\
GuideDogQA~\citep{kim2026guidedog}
& \multicolumn{1}{c}{--} & \multicolumn{1}{c}{--} & 818 & MC
& \no & \no & -- & \no & \no \\
VisAssist~\citep{gao2026visassist}
& 13,413 & 10.3 & 13,413 & OE & \yes & \yes & \no & \yes & \no \\
EgoBlind~\citep{xiao2026egoblind}
& 1,329 & 40.0 & 5,311 & OE & \yes & \yes & \yes & \yes & \no \\
\rowcolor{black!7}
\textbf{\dataset{}}
& \textbf{1,302} & \textbf{41.1} & \textbf{4,762} & \textbf{OE}
& \yes & \yes & \yes & \yes & \yes \\
\bottomrule
\end{tabular}
\Description{Comparison of nine datasets by scale, question type, and BLV-oriented annotation properties.}
\end{table}
\subsection{Dataset Statistics and Comparison}
\label{sec:data_statistics}

\dataset{} comprises 1,302 egocentric video clips and 4,762 questions across six daily assistance categories, providing 2,389 training questions (4,642 QA supervision records) and 2,373 test questions. To support standardized benchmarking while preventing overfitting, the test questions are partitioned into a standard evaluation split of 1,091 questions (2,180 accepted references, which are used across all evaluations herein) and a reserved hold-out split of 1,282 questions. As shown in Table~\ref{tab:dataset_comparison}, \dataset{} integrates authentic BLV videos, user-verified queries, time-constrained multi-reference grounding, and explicit BLV-oriented response criteria within a unified supervision and evaluation resource. These reconstructed references establish a consistent foundation for supervised fine-tuning, preference-based reinforcement learning, and BLV-centered evaluation. Comprehensive analyses of \dataset{}, including the  qualitative reconstruction comparisons and detailed distributions are provided in Appendix~A.

\begin{figure*}[t]
\centering
\includegraphics[width=\linewidth]{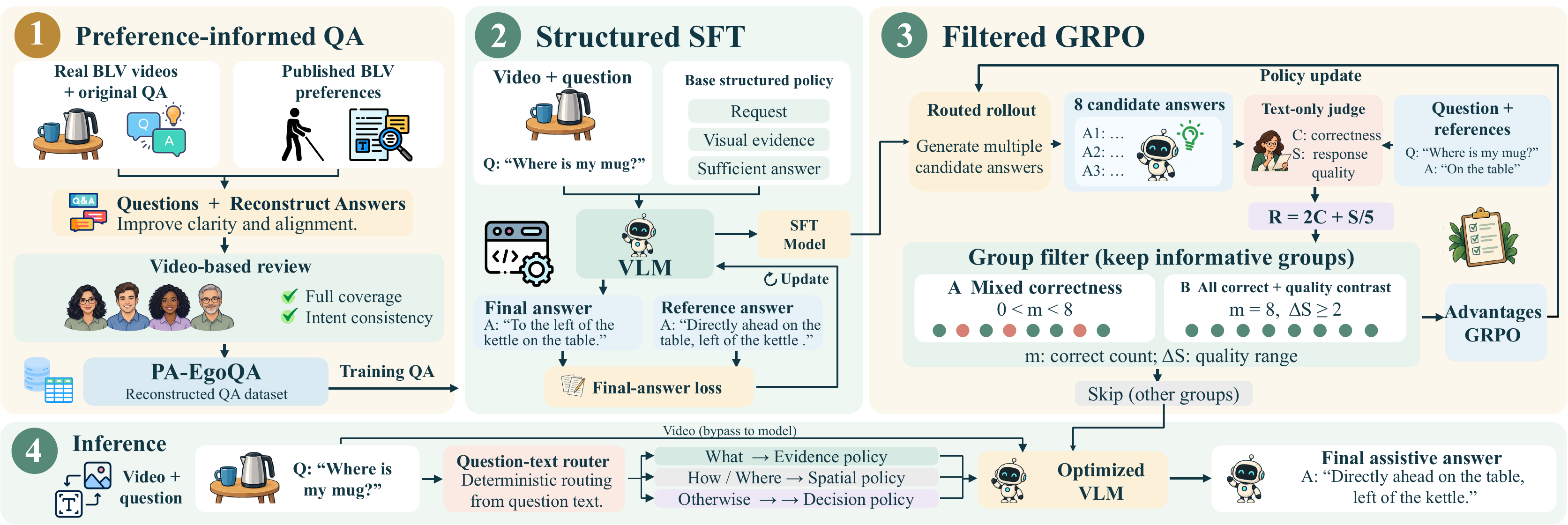}
\caption{Architecture of the proposed PAVE framework for preference-aligned egocentric learning.}
\label{fig:framework}
\end{figure*}

\section{Preference-Informed Learning and Inference}
\label{sec:method}

\subsection{Problem Formulation}
\label{sec:formulation}

Let $\mathcal{V}_q = \mathcal{F}(V_q)$ denote the visual representation extracted from the raw video segment $V_q$ via backbone-specific temporal sampling and preprocessing $\mathcal{F}(\cdot)$. Given a user query $q$ over $\mathcal{V}_q$, our objective is to learn a parameterized policy $\pi_\theta(a \mid \mathcal{V}_q, P(q))$ that generates a task-appropriate response $a = (a_1, \ldots, a_T)$ under an instruction policy $P(q)$. Crucially, we formulate egocentric visual assistance as a hierarchical multi-objective optimization problem by explicitly decoupling binary semantic correctness $C_q(a)\in\{0,1\}$ from holistic BLV assistive quality $S_q(a)\in\{0,1,\ldots,5\}$:
\begin{equation}
\label{eq:hier_obj}
\max_\theta \, \mathbb{E}_{q \sim \mathcal{D},\, a \sim \pi_\theta} \Bigl[ \mathcal{U}_{\mathrm{BLV}}\bigl(C_q(a), S_q(a)\bigr) \Bigr], \quad \text{s.t.} \quad C_q(a) \succ S_q(a),
\end{equation}
where $\mathcal{U}_{\mathrm{BLV}}(\cdot)$ denotes the lexicographic utility governed by the priority relation $\succ$, ensuring that stylistic or structural refinements never supersede factual validity.

Figure~\ref{fig:framework} illustrates the unified PAVE architecture, which integrates preference-reconstructed supervision (\dataset{}), structured answer-policy supervised fine-tuning (SFT), deterministic question-conditioned routing, and contrast-filtered group-relative policy optimization. At deployment, the optimized policy $\pi_{\theta^*}$ synthesizes a single response via deterministic single-pass greedy decoding:
\begin{equation}
\label{eq:inference}
\hat a_q = \operatorname*{arg\,max}_{a} \prod_{t=1}^{T} \pi_{\theta^*}\bigl(a_t \mid \mathcal{V}_q, P_{r(q)}(q), a_{<t}\bigr),
\end{equation}
where the active policy $P_{r(q)}$ is routed directly from the surface form of $q$.

\subsection{Structured SFT and Question-Conditioned Policy Routing}
\label{sec:structured_sft_routing}

We first fine-tune each backbone under a unified structured response policy $P_{\mathrm{str}}$ that incorporates chain-of-thought (CoT)-inspired reasoning instructions to guide task-aware answer construction. This policy enforces five operational principles: (i) resolving spatial or deictic referents in the request; (ii) grounding objects, text, and controls in visual evidence; (iii) prioritizing the final frame for current-state, directional, and safety judgments while integrating historical frames when necessary; (iv) adapting between direct factual readout and structured spatial or procedural guidance; and (v) verifying completeness while expressing uncertainty when visual evidence is inconclusive. Notably, while these instructions guide response formulation, the supervised targets contain only concise \textbf{final reference answers} without explicit intermediate reasoning traces. Given a binary answer-token mask $M_t$, SFT minimizes the normalized token-level cross-entropy:
\begin{equation}
\label{eq:sft}
\mathcal{L}_{\mathrm{SFT}}(\theta) = -\frac{\sum_{t=1}^T M_t \log \pi_\theta(a_t\mid \mathcal{V}_q, P_{\mathrm{str}}(q), a_{<t})}{\sum_{t=1}^T M_t}.
\end{equation}

Because BLV queries exhibit diverse structural demands, applying uniform CoT prompting across all questions is suboptimal: direct perception queries benefit from concise factual grounding, whereas spatial and procedural requests require structured reasoning guidance. We therefore implement three frozen, question-adaptive response policies during post-training and inference: \textbf{evidence complete answering} verifies target identity and factual details; \textbf{spatial procedural answering} emphasizes ego-relative location, temporal relations, and ordered action steps; and \textbf{decision consistent answering} prioritizes polar or choice decisions while maintaining consistency between polarity judgments and action recommendations. To eliminate routing latency, a deterministic router selects the active policy $r(q)$ directly from the normalized opening word $f(q)$, assigning $\mathrm{spatial}$ if $f(q)\in\{\mathrm{How,Where}\}$, $\mathrm{evidence}$ if $f(q)=\mathrm{What}$, and $\mathrm{decision}$ otherwise. Requiring no auxiliary classifier, this routing rule is applied identically during RL rollouts and single-pass test inference.

\subsection{Correctness-First Reward and Contrast-Filtered GRPO}
\label{sec:filter_grpo}

To align model outputs with BLV user preferences, we employ a frozen evaluator informed by prior user studies~\citep{kim2025large}. For each rollout candidate $a_{qi}$, the evaluator independently assesses binary semantic correctness $C_{qi}\in\{0,1\}$ against the accepted reference set $\mathcal{A}_q^*$ and holistic assistive quality $S_{qi}\in\{0,\ldots,5\}$ across our six BLV criteria. We combine both metrics into a hierarchical, correctness-first reward:
\begin{equation}
\label{eq:reward}
R_{qi} = 2C_{qi} + \frac{S_{qi}}{5}.
\end{equation}
Incorrect candidates receive $R_{qi}\in[0,1]$ whereas correct candidates receive $R_{qi}\in[2,3]$. Core factual correctness thus governs the primary preference ordering, preventing fluent hallucinations from outranking accurate responses, while the quality term $S_{qi}/5$ differentiates superior assistive expression among factually valid answers.

During post-training, the behavior policy $\pi_{\theta_{\mathrm{old}}}$ samples a group of $G=8$ candidate completions $\{a_{qi}\}_{i=1}^{G}$ for each query $q$ under the routed instruction $P_{r(q)}(q)$. Let $m_q = \sum_{i=1}^{G} C_{qi}$ denote the count of factually correct candidates and $\Delta S_q = \max_i S_{qi} - \min_i S_{qi}$ represent the intra-group quality span. To prevent vanishing or noisy gradients from homogeneous rollouts, we introduce a binary retention indicator $K(q)$ that selectively admits two complementary contrast regimes:
\begin{equation}
\label{eq:group_filter}
K(q) = \begin{cases}
1, & \text{if } 0 < m_q < G \quad \text{(Regime I: Correctness Contrast)}, \\
1, & \text{if } m_q = G \land \Delta S_q \ge 2 \quad \text{(Regime II: Quality Contrast)}, \\
0, & \text{otherwise}.
\end{cases}
\end{equation}
Specifically, Regime I retains groups with mixed correctness to supply discriminative gradients separating factual grounding from hallucination, whereas Regime II retains fully accurate groups with substantial quality variation to refine assistive expression. Crucially, within Regime II, the constant correctness offset ($2C_{qi}=2$) cancels out under group-mean centering, seamlessly reducing the standardized group-relative advantage $\widehat{A}_{qi}$ to a pure quality-refinement signal:
\begin{equation}
\label{eq:advantage}
\widehat{A}_{qi} = \frac{R_{qi}-\bar{R}_q}{\sigma_q+\varepsilon} = 
\begin{cases}
\dfrac{2(C_{qi}-\bar{C}_q) + \frac{1}{5}(S_{qi}-\bar{S}_q)}{\sigma_q + \varepsilon},  \text{if } 0 < m_q < G, \\[10pt]
\dfrac{S_{qi}-\bar{S}_q}{\sigma_{S,q} + 5\varepsilon},  \text{if } m_q = G \land \Delta S_q \ge 2,
\end{cases}
\end{equation}
where $\bar{R}_q$ and $\sigma_q$ denote the group mean and standard deviation of $R_{qi}$ ($\varepsilon = 10^{-4}$), and $\bar{S}_q$ and $\sigma_{S,q}$ denote the corresponding statistics of $S_{qi}$. Homogeneous groups ($K(q)=0$) yield near-zero variance and are discarded.

\begin{table*}[t]
\centering
\caption{Main results on \dataset{}. Task-wise accuracy (\%), overall accuracy (\%), and BLV Score (0--5) across twelve untuned VLMs, three matched EgoBlind baselines, and PAVE across three VLM backbones. \textbf{\#F}: sampled frames (fps: frames/s). Per-column top-5 values are shaded, with the darkest shade for the best value; the best value in each column is also bold.}
\label{tab:main_results}
\small
\setlength{\tabcolsep}{3pt}
\setlength{\aboverulesep}{0pt}
\setlength{\belowrulesep}{0pt}
\renewcommand{\arraystretch}{0.95}
\begin{tabularx}{\linewidth}{l c *{8}{Y}}
\toprule
\multirow{2}{*}{\textbf{Model / Configuration}} & \multirow{2}{*}{\textbf{\#F}} & \multicolumn{6}{c}{\textbf{Task-Category Accuracy (\%)}} & \multirow{2}{*}{\shortstack{\textbf{Overall}\\\textbf{Acc. (\%)}}} & \multirow{2}{*}{\shortstack{\textbf{BLV}\\\textbf{Score}}} \\
\cmidrule(lr){3-8}
& & \textbf{Info.} & \textbf{Safety} & \textbf{Navig.} & \textbf{Res.} & \textbf{Tool} & \textbf{Social} & & \\
\midrule
\multicolumn{10}{@{}l}{\textit{Untuned VLMs}} \\
Llama-3-VILA1.5-8B~\citep{lin2024vila} & 8 & 31.99 & 17.14 & 22.30 & 25.93 & 50.82 & 26.47 & 27.77 & 1.8955 \\
ShareGPT4Video-8B~\citep{chen2024sharegpt4video} & 16 & 34.67 & 42.86 & 18.92 & 25.93 & 37.70 & 38.24 & 34.01 & 2.2392 \\
VideoLLaMA3-7B~\citep{zhang2025videollama} & 1 fps & 41.76 & 26.53 & 31.76 & 39.51 & 31.15 & 47.06 & 36.39 & 2.2035 \\
LLaMA-VID-7B~\citep{li2024llama} & 1 fps & 40.80 & 41.22 & 20.95 & 39.51 & 19.67 & 50.00 & 37.21 & 2.2640 \\
MiniCPM-V-2.6~\citep{yao2024minicpm} & 1 fps & 46.74 & 28.57 & 26.35 & 24.69 & 52.46 & 50.00 & 38.68 & 2.2200 \\
CogVLM2-Video-Llama3-8B~\citep{hong2024cogvlm2} & 24 & 39.85 & 45.71 & 20.95 & 37.04 & 36.07 & \gThree 67.65 & 39.05 & 2.4445 \\
Video-LLaVA-7B~\citep{lin2024video} & 8 & 39.85 & 43.27 & 30.41 & 46.91 & 44.26 & 58.82 & 40.70 & 2.4335 \\
LLaVA-Video-7B-Qwen2~\citep{zhang2024llava} & 1 fps & 50.00 & 37.55 & 37.84 & 51.85 & 36.07 & 52.94 & 45.00 & 2.5885 \\
InternVL2.5-26B~\citep{chen2024expanding} & 8 & 55.75 & 38.78 & 46.62 & 38.27 & \gThree 70.49 & 50.00 & 50.05 & 2.7617 \\
Qwen2.5-VL-7B-Instruct~\citep{bai2025qwen25vltechnicalreport} & 1 fps & 53.83 & 42.04 & 43.92 & 51.85 & \gFour 72.13 & 58.82 & 50.87 & 2.7259 \\
LLaVA-OneVision-Qwen2-7B~\citep{li2024llava} & 16 & 54.79 & 42.86 & 42.57 & 51.85 & \gTwo 63.93 & \gOne 61.76 & 50.96 & 2.8524 \\
InternVL2.5-8B~\citep{chen2024expanding} & 8 & 56.32 & 46.53 & \gOne 52.03 & 51.85 & \gFive \textbf{75.41} & 58.82 & 54.35 & \gOne 2.9074 \\
\midrule
\multicolumn{10}{@{}l}{\textit{EgoBlind}} \\
Qwen2.5-VL-7B & 1 fps & \gOne 57.28 & 47.35 & \gTwo 54.73 & \gTwo 58.02 & 55.74 & \gOne 61.76 & 54.81 & 2.8588 \\
LLaVA-OV-7B & 16 & 55.36 & \gTwo 65.31 & \gFour 64.86 & \gTwo 58.02 & 54.10 & \gFour 70.59 & \gTwo 59.49 & \gTwo 3.0147 \\
InternVL2.5-8B & 8 & \gFour 61.69 & \gFour 69.39 & \gThree 57.43 & \gThree 59.26 & 60.66 & \gFour 70.59 & \gThree 62.88 & \gThree 3.0660 \\
\midrule
\multicolumn{10}{@{}l}{\textbf{\textit{PAVE (Ours)}}} \\
Qwen2.5-VL-7B & 1 fps & \gThree 59.39 & \gOne 52.24 & 50.68 & \gOne 53.09 & 49.18 & 61.76 & \gOne 55.64 & 2.8478 \\
LLaVA-OV-7B & 16 & \gTwo 59.20 & \gThree 66.53 & \gFive \textbf{70.27} & \gFive \textbf{67.90} & \gThree 70.49 & \gTwo 64.71 & \gFour 63.79 & \gFour 3.1210 \\
InternVL2.5-8B & 8 & \gFive \textbf{64.56} & \gFive \textbf{75.10} & \gFour 64.86 & \gFour 62.96 & \gOne 62.30 & \gFive \textbf{73.53} & \gFive \textbf{67.00} & \gFive \textbf{3.2878} \\
\bottomrule
\end{tabularx}
\Description{The table compares six task-category accuracies and the overall accuracy and BLV Score for twelve untuned VLMs, three matched EgoBlind models, and three PAVE models. PAVE on InternVL achieves 67.00 percent accuracy and 3.2878 BLV Score.}
\end{table*}

Broadcasting $\widehat{A}_{qi}$ across all tokens $t \in \{1,\ldots,T_{qi}\}$ of completion $a_{qi}$, let $\rho_{qit}(\theta)$ denote the token-level importance sampling ratio between $\pi_\theta$ and $\pi_{\theta_{\mathrm{old}}}$, and let $\mathbb{D}_{\mathrm{KL}}(\pi_\theta \,\|\, \pi_{\mathrm{ref}})_{qit}$ denote the unbiased Schulman KL divergence estimator relative to the frozen structured-SFT reference policy $\pi_{\mathrm{ref}}$. The clipped per-token objective~\citep{shao2024deepseekmath} is defined as:
\begin{align}
\label{eq:grpo_token}
\ell_{qit}(\theta) ={}& -\min\Bigl(\rho_{qit}(\theta)\widehat{A}_{qi},\, \operatorname{clip}\bigl(\rho_{qit}(\theta), 1-\epsilon, 1+\epsilon\bigr)\widehat{A}_{qi}\Bigr) \nonumber \\
& + \beta\,\mathbb{D}_{\mathrm{KL}}(\pi_\theta \,\|\, \pi_{\mathrm{ref}})_{qit},
\end{align}
with clipping threshold $\epsilon=0.2$ and KL penalty coefficient $\beta=0.04$. Accumulating retained groups into an effective batch $\mathcal{B}$ ($|\mathcal{B}|=2G=16$ completions), the overall \filter{} objective averages token losses per sequence prior to batch-level aggregation:
\begin{equation}
\label{eq:grpo_full}
\mathcal{L}_{\mathrm{FGRPO}}(\theta) = \frac{1}{|\mathcal{B}|}\sum_{(q,i)\in\mathcal{B}:\, K(q)=1} \frac{1}{T_{qi}}\sum_{t=1}^{T_{qi}} \ell_{qit}(\theta).
\end{equation}
Across InternVL2.5-8B, LLaVA-OneVision-7B, and Qwen2.5-VL-7B, \filter{} freezes the vision encoder and structured-SFT weights ($\pi_{\mathrm{ref}}$), updating only newly attached rank-16 LoRA adapters ($\alpha=16$, dropout $0$) on the language linear projections and multimodal connector.

\section{Experiments}
\label{sec:experiments}

\subsection{Experimental Setup}
\label{sec:experiment_setup}
\textbf{Implementation details.} We follow the native visual preprocessing of each backbone: InternVL2.5-8B samples 8 frames, LLaVA-OV-7B samples 16 frames, and Qwen2.5-VL-7B uses 1-fps sampling. Structured SFT follows standard backbone adaptation, whereas \filter{} updates rank-16 LoRA adapters on the language projections and multimodal connector using $G=8$ rollouts. More details are provided in Appendix~\ref{app:implementation}.

\textbf{Dataset and backbones.} All task-adapted models are trained and evaluated on \dataset{}. Our primary investigation encompasses three open-weights VLM backbones: InternVL2.5-8B~\citep{chen2024expanding}, LLaVA-OneVision-7B~\citep{li2024llava}, and Qwen2.5-VL-7B~\citep{bai2025qwen25vltechnicalreport}, alongside twelve zero-shot VLMs~\citep{lin2024vila,chen2024sharegpt4video,zhang2025videollama,li2024llama,yao2024minicpm,hong2024cogvlm2,lin2024video,zhang2024llava,chen2024expanding,bai2025qwen25vltechnicalreport,li2024llava}.

\textbf{Baselines and metrics.} We compare PAVE against: (1) Zero-Shot VLMs,
comprising twelve off-the-shelf models evaluated without task-specific tuning,
and (2) EgoBlind, a matched baseline fine-tuned under the original EgoBlind
protocol and prompt~\citep{xiao2026egoblind} using \dataset{} supervision.
Using a frozen, reference-grounded evaluator, we report decoupled factual
Accuracy and holistic BLV Score (scaled 0--5) across $N$ test queries:
\begin{equation}
  \mathrm{Acc.} = \frac{100}{N}\sum_{q=1}^{N} C_q,
  \mathrm{BLV}  = \frac{1}{N}\sum_{q=1}^{N} S_q .
\label{eq:metrics}
\end{equation}

\subsection{Quantitative Evaluation}
\label{sec:main_results}

\textbf{Overall performance across backbones.} Table~\ref{tab:main_results} shows that PAVE consistently outperforms matched EgoBlind baselines across all three backbones. On InternVL2.5-8B, PAVE attains 67.00\% accuracy (+4.12\%) and a 3.2878 BLV Score (+0.2218). Similarly, LLaVA-OV-7B reaches 63.79\% accuracy (+4.30\%) and a 3.1210 BLV Score (+0.1063), while Qwen2.5-VL-7B reaches 55.64\% accuracy (+0.83\%) with comparable assistive quality ($\approx 2.85$). Notably, all PAVE models surpass the strongest zero-shot baseline (InternVL2.5-8B, 54.35\%), improving accuracy by +12.65\% on InternVL2.5-8B and +12.83\% on LLaVA-OV-7B over their zero-shot counterparts.

\begin{figure}[t]
\centering
\vspace{-2.5pt}
\includegraphics[width=\columnwidth]{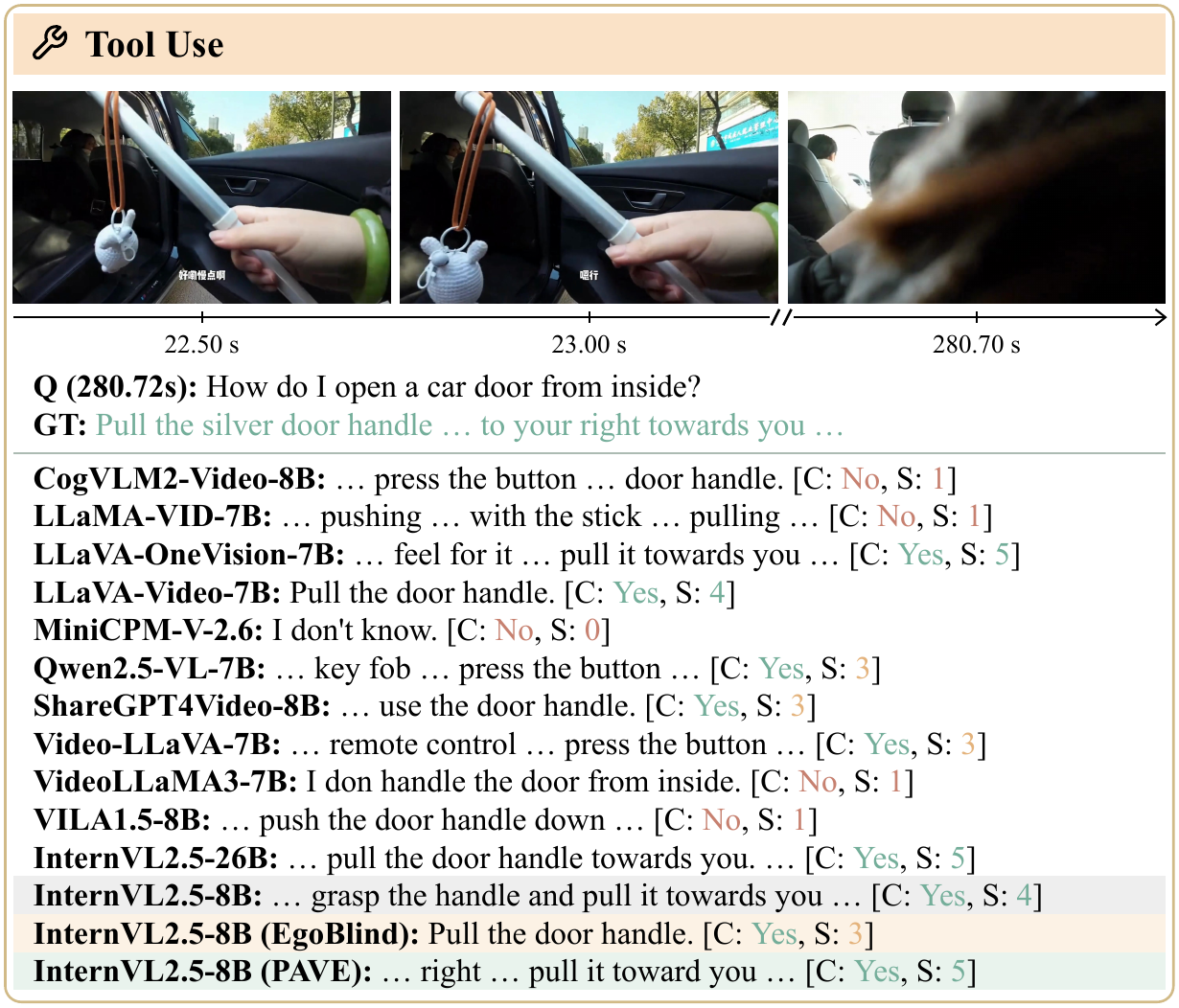}
\caption{Qualitative visualization on a situated \emph{Tool Use} query. PAVE resolves ego-relative orientation and manipulation actions ($S_q=3\to 5$).}
\label{fig:main_tool_case}
\vspace{-12pt}
\end{figure}

\textbf{Task-category performance.} Across assistance categories, the InternVL2.5-8B improves Navigation accuracy by +7.43\% and reaches 75.10\% on Safety Warnings, whereas LLaVA-OV-7B leads on Navigation and Other Resources (Table~\ref{tab:main_results}). Qwen2.5-VL-7B improves Safety Warnings and Information Reading by +4.89\% and +2.11\% over EgoBlind. Standard fine-tuning (EgoBlind) struggles on spatial tasks because it imitates viewer-agnostic descriptions. By penalizing vague deictic references via its hierarchical reward, PAVE translates visual cues into precise, ego-relative guidance. Furthermore, while zero-shot VLMs occasionally score higher raw accuracy on \emph{Tool Use} by listing speculative actions in verbose outputs (guessing both buttons and handles triggers $C_q=1$ despite low utility; Figure~\ref{fig:main_tool_case}), PAVE enforces concise, hallucination-free guidance, boosting Tool Use accuracy by +16.39\% on LLaVA-OV-7B and +1.64\% on InternVL2.5-8B alongside substantial BLV Score gains.

\begin{table}[t]
\centering
\caption{Accuracy (\%) breakdown by question form on InternVL2.5-8B. Parentheses denote PAVE gains.}
\label{tab:app_question_forms}
\small
\setlength{\tabcolsep}{3pt}
\setlength{\aboverulesep}{0pt}
\setlength{\belowrulesep}{0pt}
\renewcommand{\arraystretch}{1.0}
\providecommand{\qformdelta}[1]{{\scriptsize\textcolor{gradBase!80!black}{\,(#1)}}}
\begin{tabularx}{\linewidth}{l *{3}{Y}}
\toprule
\textbf{Question Form} & \textbf{Zero-Shot} & \textbf{EgoBlind} & \textbf{PAVE (Ours)} \\
\midrule
What & 44.93 \qformdelta{+4.85\%} & \underline{49.34} \qformdelta{+0.44\%} & \textbf{49.78} \\
Where & 38.46 \qformdelta{+11.54\%} & \textbf{50.00} & \textbf{50.00} \\
How & \underline{60.17} \qformdelta{+3.39\%} & 52.54 \qformdelta{+11.02\%} & \textbf{63.56} \\
Which & 56.86 \qformdelta{+9.81\%} & \underline{64.71} \qformdelta{+1.96\%} & \textbf{66.67} \\
Other & \underline{68.00} \qformdelta{+4.00\%} & 64.00 \qformdelta{+8.00\%} & \textbf{72.00} \\
Yes/No & 59.01 \qformdelta{+18.55\%} & \underline{72.61} \qformdelta{+4.95\%} & \textbf{77.56} \\
\midrule
\textbf{Overall} & 54.35 \qformdelta{+12.65\%} & \underline{62.88} \qformdelta{+4.12\%} & \textbf{67.00} \\
\bottomrule
\end{tabularx}
\end{table}

\begin{table}[t]
\centering
\caption{Response economy (mean generated tokens) and end-to-end single-pass inference latency (seconds per query).}
\label{tab:response_economy}
\small
\setlength{\tabcolsep}{4pt}
\setlength{\aboverulesep}{0pt}
\setlength{\belowrulesep}{0pt}
\renewcommand{\arraystretch}{1.0}
\begin{tabularx}{\linewidth}{l l *{2}{Y}}
\toprule
\textbf{Backbone} & \textbf{Configuration} & \textbf{Tokens $\downarrow$} & \textbf{Latency (s) $\downarrow$} \\
\midrule
\multirow{2}{*}{Qwen} & EgoBlind & 14.71 & 6.97 \\
& \textbf{PAVE (Ours)} & \textbf{7.23} & \textbf{6.88} \\
\midrule
\multirow{2}{*}{LLaVA} & EgoBlind & 11.64 & \textbf{4.45} \\
& \textbf{PAVE (Ours)} & \textbf{11.63} & 4.48 \\
\midrule
\multirow{2}{*}{InternVL} & EgoBlind & 10.36 & 2.66 \\
& \textbf{PAVE (Ours)} & \textbf{6.44} & \textbf{2.52} \\
\bottomrule
\end{tabularx}
\Description{Mean generated token counts and end-to-end inference latency comparing EgoBlind and PAVE.}
\end{table}

\textbf{Fine-grained question-form evaluation.}
Table~\ref{tab:app_question_forms} partitions the evaluation queries on InternVL2.5-8B by their opening word. PAVE achieves its largest gains over EgoBlind on \emph{How} (+11.02\%), \emph{Other} (+8.00\%), and \emph{Yes/No} (+4.95\%). On procedural \emph{How} and open-ended \emph{Other} queries, static fine-tuning under a single uniform prompt (EgoBlind) causes severe adaptation degradation below the zero-shot model (e.g., \emph{How} accuracy drops from 60.17\% to 52.54\%). Conversely, our question-conditioned routing preserves concise grounding on direct perception queries while dynamically activating spatial-procedural guidance for complex requests, reversing this degradation to surpass the zero-shot baseline (63.56\% on \emph{How} and 72.00\% on \emph{Other}). Assistive quality also improves across all six forms, led by BLV Score gains on \emph{How} (+0.3475) and \emph{Yes/No} (+0.2809). On \emph{Where} queries, PAVE improves accuracy by +11.54\% over the zero-shot model (matching the 50.00\% peak) while elevating the BLV Score by +0.1153 over EgoBlind, showing how regime-shifted quality optimization refines spatial expression even at saturated accuracy (full results in Appendix~\ref{app:question_form_full}).

\textbf{Inference efficiency and response economy.}
For BLV users relying on text-to-speech readout, concise responses and low latency are critical. As shown in Table~\ref{tab:response_economy}, PAVE compresses mean output length by 50.8\% on Qwen2.5-VL-7B (14.71 to 7.23 tokens) and 37.8\% on InternVL2.5-8B (10.36 to 6.44 tokens), while maintaining the compact 11.6-token footprint and 4.48~s latency of LLaVA-OV-7B. End-to-end per-query latency decreases from 6.97~s to 6.88~s on Qwen2.5-VL-7B and from 2.66~s to 2.52~s on InternVL2.5-8B. Unlike conventional CoT methods that generate verbose intermediate rationales, PAVE internalizes structured verification during training while supervising solely on concise final answers, delivering high-utility guidance at up to half the token cost.

\begin{table}[t]
\centering
\caption{Ablation study on preference-reconstructed supervision (\dataset{}) versus original EgoBlind annotations.}
\label{tab:matched_supervision}
\small
\setlength{\tabcolsep}{4pt}
\setlength{\aboverulesep}{0pt}
\setlength{\belowrulesep}{0pt}
\renewcommand{\arraystretch}{1.0}
\providecommand{\supDelta}[1]{{\scriptsize\textcolor{gradBase!80!black}{\,(#1)}}}
\begin{tabularx}{\linewidth}{l l *{2}{Y}}
\toprule
\textbf{Backbone} & \textbf{Supervision Target} & \textbf{Acc. (\%)} & \textbf{BLV Score} \\
\midrule
\multirow{2}{*}{Qwen} & Original (EgoBlind) & 52.15 & 2.8579 \\
& \textbf{\dataset{} (Ours)} & \textbf{54.81}\supDelta{+2.66\%} & \textbf{2.8588}\supDelta{+0.0009} \\
\midrule
\multirow{2}{*}{InternVL} & Original (EgoBlind) & 55.64 & 2.9157 \\
& \textbf{\dataset{} (Ours)} & \textbf{62.88}\supDelta{+7.24\%} & \textbf{3.0660}\supDelta{+0.1503} \\
\midrule
\multirow{2}{*}{LLaVA} & Original (EgoBlind) & 53.07 & 2.7883 \\
& \textbf{\dataset{} (Ours)} & \textbf{59.49}\supDelta{+6.42\%} & \textbf{3.0147}\supDelta{+0.2264} \\
\bottomrule
\end{tabularx}
\Description{Comparison between original EgoBlind annotations and PA-EgoQA supervision across Qwen, InternVL, and LLaVA.}
\end{table}

\begin{table}[t]
\centering
\caption{Ablation study on contrast-filtered policy optimization under identical question-conditioned routing.}
\label{tab:grpo_ablation}
\small
\setlength{\tabcolsep}{3pt}
\setlength{\aboverulesep}{0pt}
\setlength{\belowrulesep}{0pt}
\renewcommand{\arraystretch}{1.0}
\providecommand{\supDelta}[1]{{\scriptsize\textcolor{gradBase!80!black}{\,(#1)}}}
\begin{tabularx}{\linewidth}{l l *{2}{Y}}
\toprule
\textbf{Backbone} & \textbf{Configuration} & \textbf{Acc. (\%)} & \textbf{BLV Score} \\
\midrule
\multirow{2}{*}{Qwen} & Structured SFT + Routing & 55.27 & \textbf{2.8497} \\
& \textbf{PAVE (Ours)} & \textbf{55.64}\supDelta{+0.37\%} & 2.8478\supDelta{$-$0.0019} \\
\midrule
\multirow{2}{*}{LLaVA} & Structured SFT + Routing & 63.06 & 3.1072 \\
& \textbf{PAVE (Ours)} & \textbf{63.79}\supDelta{+0.73\%} & \textbf{3.1210}\supDelta{+0.0138} \\
\midrule
\multirow{2}{*}{InternVL} & Structured SFT + Routing & 66.09 & 3.1934 \\
& \textbf{PAVE (Ours)} & \textbf{67.00}\supDelta{+0.91\%} & \textbf{3.2878}\supDelta{+0.0944} \\
\bottomrule
\end{tabularx}
\Description{Structured SFT plus routing and full PAVE are compared for each of three VLM backbones.}
\end{table}

\subsection{Ablation Studies}
\label{sec:component_results}
\label{sec:question_forms}
\label{sec:visual_controls}
\label{sec:response_economy}

\textbf{Effect of preference-reconstructed supervision.} Isolating reference reconstruction under identical training questions and EgoBlind prompts (Table~\ref{tab:matched_supervision}), substituting original annotations with \dataset{} references consistently improves accuracy by +2.66\% on Qwen2.5-VL-7B, +7.24\% on InternVL2.5-8B, and +6.42\% on LLaVA-OV-7B, alongside BLV Score gains of +0.0009, +0.1503, and +0.2264. This confirms that embedding BLV communication criteria into supervision targets directly enhances situated assistance.

\textbf{Effect of contrast-filtered policy optimization.} Under identical routing (Table~\ref{tab:grpo_ablation}), adding \filter{} post-training further elevates accuracy across all backbones (+0.91\% on InternVL2.5-8B, +0.73\% on LLaVA-OV-7B, and +0.37\% on Qwen2.5-VL-7B). It also resulted in an improvement in BLV scores, validating the efficacy of dual-regime contrast filtering.

\textbf{Effect of question-conditioned policy routing.} Table~\ref{tab:policy_controls} compares Direct QA, Uniform CoT, and our deterministic router under Structured SFT. Relative to Direct QA, routing improves accuracy by +3.58\% on InternVL2.5-8B and +2.75\% on LLaVA-OV-7B. While Uniform CoT degrades accuracy on Qwen2.5-VL-7B (53.07\% vs.\ 55.55\% under Direct QA) by imposing complex rationales on simple perception queries, our router shields direct perception requests from reasoning overhead. This targeted allocation restores accuracy on Qwen2.5-VL-7B to 55.27\% and drives peak SFT accuracy on LLaVA-OV-7B (+3.11\%) and InternVL2.5-8B (+1.10\%).
\begin{table}[!tbp]
\centering
\caption{Ablation study on inference-time response policies under Structured
SFT. Bold and underline mark the best and second-best result within each
backbone block, respectively.}
\label{tab:policy_controls}
\small
\setlength{\tabcolsep}{3pt}
\setlength{\aboverulesep}{0pt}
\setlength{\belowrulesep}{0pt}
\renewcommand{\arraystretch}{1.0}
\begin{tabularx}{\linewidth}{l l *{2}{Y}}
\toprule
\textbf{Backbone} & \textbf{Inference Policy} & \textbf{Acc. (\%)} & \textbf{BLV Score} \\
\midrule
\multirow{3}{*}{Qwen} & Direct QA & \textbf{55.55} & \textbf{2.9551} \\
& Uniform CoT & 53.07 & 2.8442 \\
& \textbf{Question-Routed} & \underline{55.27} & \underline{2.8497} \\
\midrule
\multirow{3}{*}{LLaVA} & Direct QA & \underline{60.31} & \underline{3.0797} \\
& Uniform CoT & 59.95 & 3.0541 \\
& \textbf{Question-Routed} & \textbf{63.06} & \textbf{3.1072} \\
\midrule
\multirow{3}{*}{InternVL} & Direct QA & 62.51 & 3.1430 \\
& Uniform CoT & \underline{64.99} & \textbf{3.2548} \\
& \textbf{Question-Routed} & \textbf{66.09} & \underline{3.1934} \\
\bottomrule
\end{tabularx}
\Description{Direct QA, uniform CoT reasoning, and question-conditioned routing are compared under Structured SFT.}
\end{table}

\begin{table}[!tbp]
\centering
\caption{Ablation study on visual grounding and temporal context. Acc. denotes
Accuracy (\%) and BLV denotes BLV Score. Bold and underline mark the best and
second-best result within each column, respectively.}
\label{tab:visual_controls}
\small
\setlength{\aboverulesep}{0pt}
\setlength{\belowrulesep}{0pt}
\renewcommand{\arraystretch}{1.1}
\begin{tabularx}{\linewidth}{l *{6}{Y}}
\toprule
\multirow{2}{*}{\textbf{Condition}}
& \multicolumn{2}{c}{\textbf{InternVL}}
& \multicolumn{2}{c}{\textbf{LLaVA}}
& \multicolumn{2}{c}{\textbf{Qwen}} \\
\cmidrule(lr){2-3}\cmidrule(lr){4-5}\cmidrule(lr){6-7}
& \textbf{Acc.} & \textbf{BLV}
& \textbf{Acc.} & \textbf{BLV}
& \textbf{Acc.} & \textbf{BLV} \\
\midrule
Zero-Frame & 27.04 & 1.6453 & 27.13 & 1.7791 & 30.98 & 1.9633 \\
Last-Frame & 59.30 & 3.0183 & 58.39 & 3.0037 & \textbf{55.36} & \textbf{2.9056} \\
Shuffle-Frame & \underline{64.62} & \textbf{3.1943} & \underline{62.88} & \underline{3.1063} & 52.15 & 2.7892 \\
\textbf{Full Video} & \textbf{66.09} & \underline{3.1934} & \textbf{63.06} & \textbf{3.1072} & \underline{55.27} & \underline{2.8497} \\
\bottomrule
\end{tabularx}
\end{table}

\textbf{Effect of visual grounding and temporal context.} Table~\ref{tab:visual_controls} evaluates visual and temporal reliance across four input conditions. Removing visual input (Zero-Frame) reduces accuracy by 24\%--39\%, confirming strict reliance on visual evidence. Incorporating historical frames (Full Video) improves accuracy over Last-Frame by +6.79\% on InternVL2.5-8B and +4.67\% on LLaVA-OV-7B. Under the fixed total pixel budget of Qwen2.5-VL-7B, dedicating all spatial resolution to the final frame (55.36\%) trades temporal coverage for spatial detail, matching multi-frame Full Video (55.27\%). Crucially, under identical multi-frame budgets, Full Video consistently outperforms Shuffle-Frame across all backbones (e.g., +3.12\% on Qwen2.5-VL-7B), confirming that PAVE exploits coherent temporal progression rather than merely aggregating  static frames.

\begin{figure}[t]
\centering
\includegraphics[width=0.98\columnwidth]{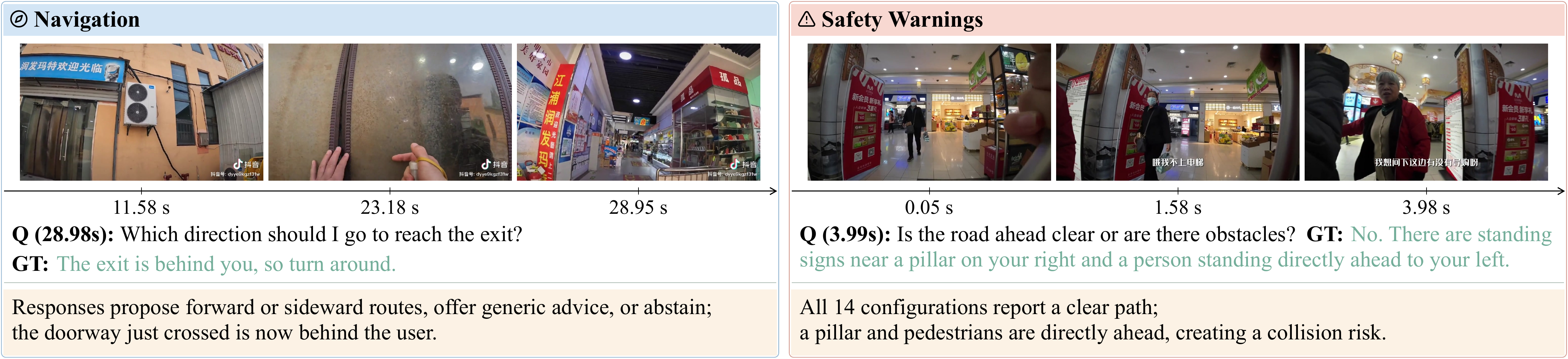}
\vspace{-6pt}
\caption{Failure cases requiring spatial updating across turns (left) and hazard perception under motion blur (right).}
\label{fig:main_failure_cases}
\vspace{-12pt}
\end{figure}

\subsection{Qualitative and Visualization Analysis}
\label{sec:qualitative_main}

\textbf{Case visualization on situated queries.} Figure~\ref{fig:main_tool_case} visualizes model predictions on a situated manipulation query, illustrating how PAVE improves factual grounding and assistive utility. Zero-shot VLMs frequently hallucinate nonexistent buttons. While EgoBlind identifies the target (\emph{``Pull the door handle''}, $C_q=1$), it omits spatial orientation ($S_q=3$). PAVE specifies both the ego-relative location (\emph{``to your right''}) and action (\emph{``pull it toward you''}), achieving $S_q=5$ (more details are provided in Appendix~\ref{app:qualitative_responses}).

\textbf{Failure case analysis and bottlenecks.} Figure~\ref{fig:main_failure_cases} visualizes two challenging scenarios where all evaluated models fail: updating the direction of an exit after a large viewpoint shift, and detecting hazards under severe motion blur. Thus, while PAVE aligns responses with BLV preferences, rapid ego-motion tracking and blurred hazard perception are constrained by the native VLM backbones.

\section{Conclusion}
\label{sec:conclusion}
In this work, we address a fundamental discrepancy in egocentric visual assistance for blind and low-vision users: semantic agreement with reference answers does not guarantee situated assistive utility. To bridge this gap, we construct \dataset{}, a preference-aligned benchmark that reconstructs supervision around blind empirical communication criteria. Building upon this foundation, we introduce PAVE, a unified learning and single-pass inference framework that explicitly decouples core factual validity from holistic assistive quality. By integrating structured answer-policy supervision, zero-overhead deterministic question-conditioned routing, and contrast-filtered policy optimization, PAVE adapts response formulation to the structural demands of diverse queries while learning from both factual accuracy contrasts and fine-grained quality variations. Extensive quantitative, ablation, and qualitative evaluations across multiple vision-language backbones confirm that our framework simultaneously enhances factual grounding, elevates ego-relative assistive expression, and improves response conciseness without additional inference latency. Ultimately, our work establishes a principled paradigm for human-centered blind and low-vision accessibility requirements.

\appendix

\section{\dataset{} Dataset Analysis}
\label{app:dataset_analysis}

\subsection{Original Reference Quality Audit}
\label{app:original_quality}

\begin{figure}[!htbp]
\centering
\includegraphics[
    width=\columnwidth,
    height=0.28\textheight,
    keepaspectratio
]{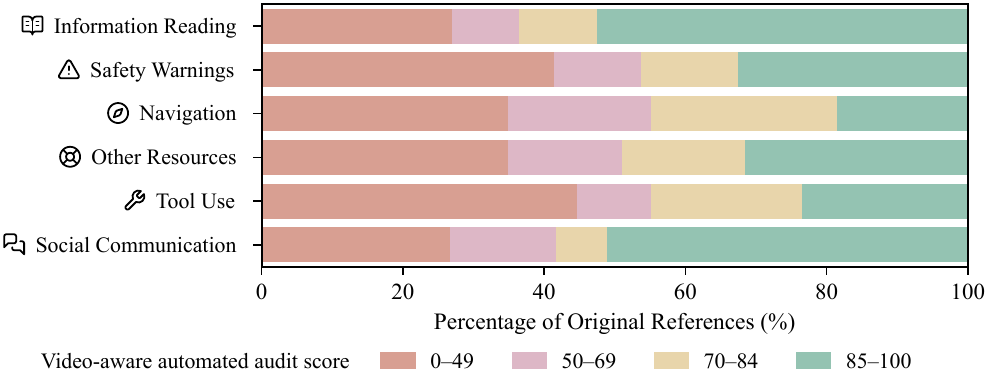}
\caption{Distribution of video-grounded audit scores ($J_{\mathrm{vid}}$) for original EgoBlind references across assistance categories.}
\Description{Six stacked bars compare reference-answer audit score distributions across four intervals: 0--49, 50--69, 70--84, and 85--100.}
\label{fig:app_original_quality}
\end{figure}

Our core premise in Section~\ref{sec:reconstruction} is that semantic agreement with raw annotations does not guarantee situated assistive utility. Figure~\ref{fig:app_original_quality} supports this premise: it shows how video-grounded audit scores ($J_{\mathrm{vid}}\in[0,100]$) are distributed across all six assistance categories in the original EgoBlind annotations, before
reconstruction. While \emph{Information Reading} exhibits the highest proportion of original references satisfying our strict acceptance threshold ($\tau=85$), spatially and procedurally demanding categories—specifically \emph{Navigation}, \emph{Tool Use}, and \emph{Safety Warnings}—frequently fail to meet this standard due to the omission of ego-relative orientation, hazard specificity, or actionable granularity. This empirical discrepancy directly motivates our preference-informed reference reconstruction in \dataset{}.

\subsection{Preference-Informed Reference Examples}
\label{app:reference_examples}

\begin{figure*}[!t]
\centering
\includegraphics[width=0.9\textwidth]{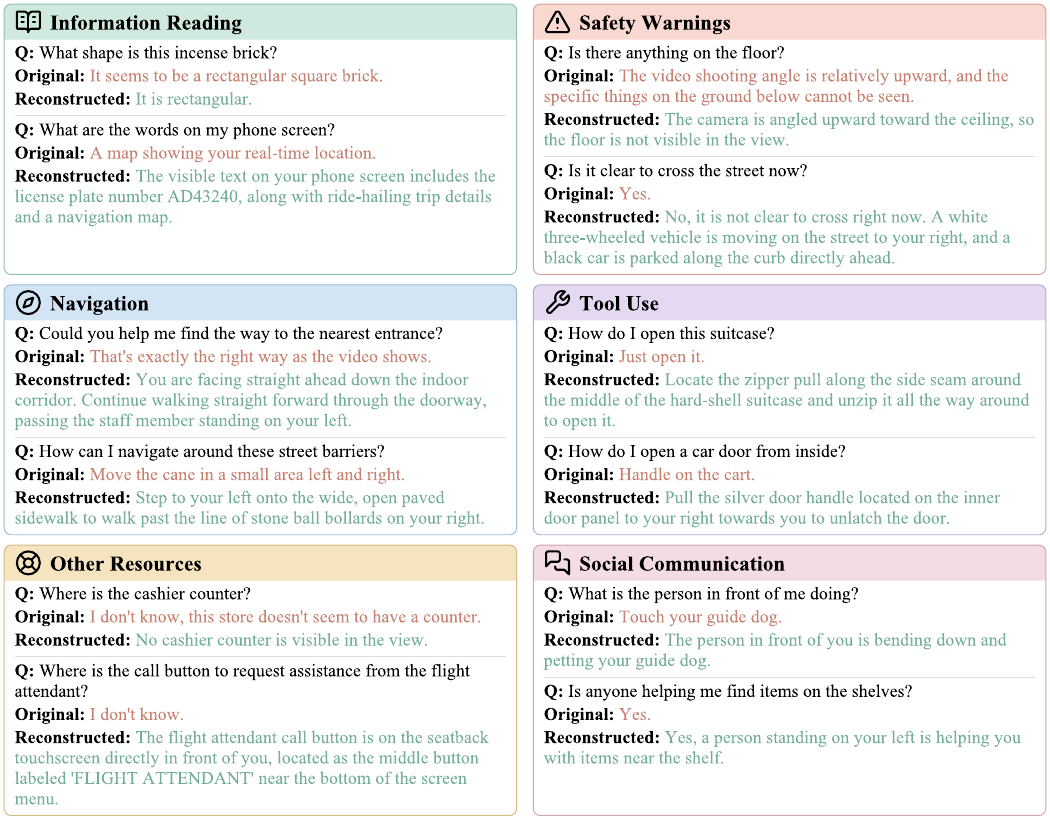}
\caption{Qualitative comparison between original EgoBlind annotations and reconstructed \dataset{} references across all six assistance categories.}
\label{fig:app_reference_examples}
\Description{Six task panels each compare original and reconstructed answers to two BLV questions.}
\end{figure*}

Figure~\ref{fig:app_reference_examples} presents twelve representative pairs comparing original EgoBlind annotations against reconstructed \dataset{} references across all six assistance categories, illustrating how our multi-stage reconstruction pipeline and four-reviewer video audit operationalize the principle of \textbf{minimum sufficient assistance} (Section~\ref{sec:reconstruction_criteria}). Specifically, the reconstructed targets establish explicit ego-relative spatial anchoring, decompose complex appliance manipulations into sequential action steps, ground claims strictly in visible evidence, and prioritize immediate physical hazards, while preserving concise phrasing whenever a brief factual readout suffices.

\subsection{Linguistic and Temporal Characteristics}
\label{app:linguistic_temporal}

\begin{figure*}[!t]
\centering
\includegraphics[width=0.87\textwidth]{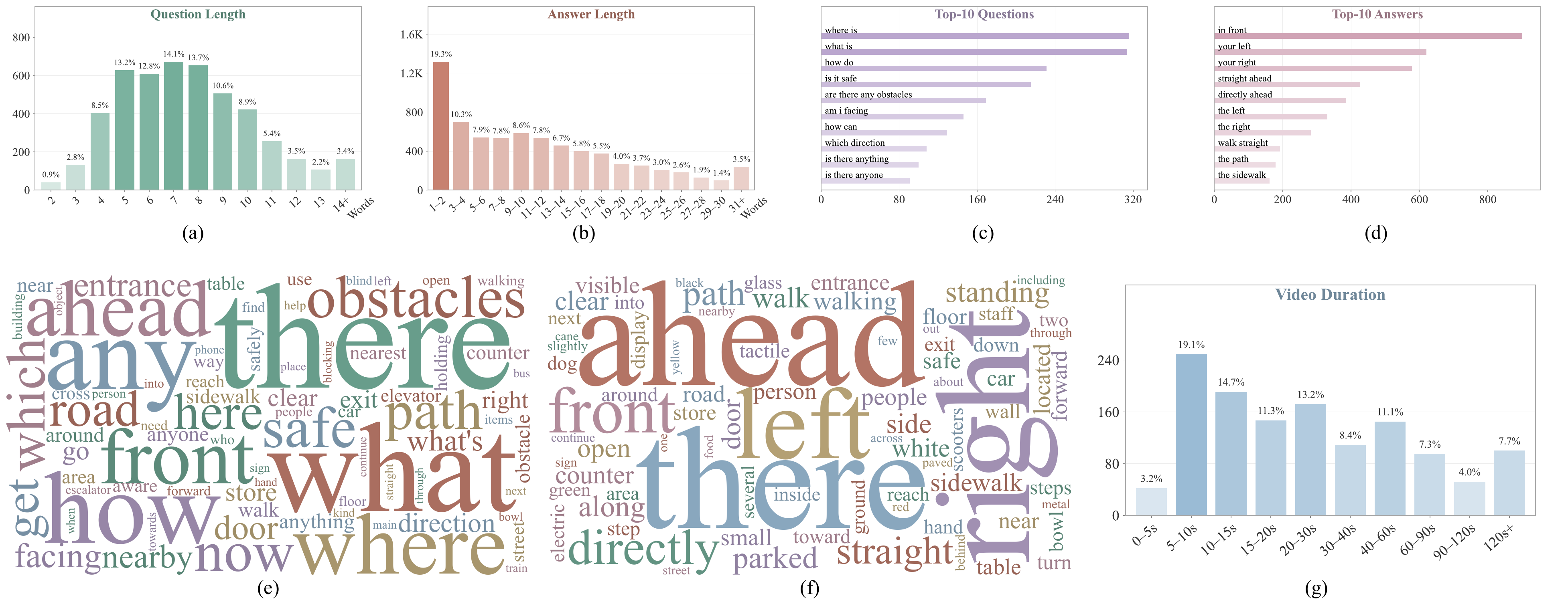}
\caption{Linguistic and temporal statistics of \dataset{}: (a, b) question and reference length distributions; (c, d) top-frequency $n$-gram expressions; (e, f) vocabulary word clouds; and (g) video clip duration distribution.}
\label{fig:app_dataset_statistics}
\Description{Seven plots depict question and reference-answer lengths, frequent question and answer expressions, word clouds, and video durations.}
\end{figure*}

Figure~\ref{fig:app_dataset_statistics} characterizes the linguistic and temporal distributions of \dataset{}. The query-level analyses encompass all 4,762 retained questions across 1,302 unique egocentric video clips, whereas the reference-level analyses cover all 6,822 accepted reference answers spanning the training (4,642) and standard evaluation (2,180) splits. The diverse query openings reflect authentic, situated daily assistance requests, providing the empirical foundation for our deterministic question-conditioned policy router (Section~\ref{sec:structured_sft_routing}). Concurrently, the length distributions and high-frequency $n$-gram expressions in the reconstructed references underscore compact, user-relative spatial anchoring and actionable guidance.

\section{Supplementary Experimental Results}
\label{app:supplementary_results}
\label{app:extended_experiments}

\subsection{Full Question-Form Evaluation across Backbones}
\label{app:question_form_full}

Table~\ref{tab:app_question_form_full} reports the complete breakdown of factual Accuracy and holistic BLV Score across canonical question forms for all three VLM backbones on the 1,091 evaluation queries, complementing the InternVL2.5-8B analysis in Table~3. Consistent with the main evaluation in Section~5, we compare PAVE against off-the-shelf \textbf{Zero-Shot} baselines and matched \textbf{EgoBlind} fine-tuned baselines. Beyond validating the effectiveness of deterministic question-conditioned routing, these results reflect architectural differences in visual token allocation: whereas InternVL2.5-8B and LLaVA-OV-7B (which preserve high per-frame spatial resolution across sampled frames) achieve substantial gains on multi-step spatial and procedural queries (\emph{How} and \emph{Where}), the shared 1-fps pixel budget and 50.8\% token compression of Qwen2.5-VL-7B concentrate its improvements on high-density decision, safety verification, and reading queries (\emph{Which}, \emph{Yes/No}, and \emph{Other}).

\textbf{Question-form-specific gains and regime-shifted refinement.}
PAVE consistently elevates both Accuracy and BLV Score on \emph{Which} and \emph{Yes/No} queries across all three backbones, confirming the efficacy of our \textbf{decision-consistent answering} policy ($P_{\mathrm{decision}}$). Substantial improvements also emerge on procedural \emph{How} queries governed by our \textbf{spatial-procedural answering} policy ($P_{\mathrm{spatial}}$): relative to EgoBlind, InternVL2.5-8B improves Accuracy by +11.02\% and BLV Score by +0.3475, while LLaVA-OV-7B achieves corresponding gains of +8.47\% and +0.2542. On \emph{Which} queries, LLaVA-OV-7B increases Accuracy by +17.65\% and BLV Score by +0.4314, while Qwen2.5-VL-7B achieves gains of +7.84\% and +0.3333, respectively. Crucially, on \emph{Where} queries under InternVL2.5-8B, PAVE matches the 50.00\% accuracy ceiling of EgoBlind while elevating the BLV Score from 2.5385 to 2.6538 (+0.1153), directly demonstrating how Regime~II quality contrast in \filter{} (Equation~8) refines spatial expression even when factual correctness saturates.

\begin{table*}[!t]
\centering
\caption{Complete question-form breakdown on \dataset{} across three VLM backbones. Factual Accuracy (\%) and holistic BLV Score (0--5) are reported for Zero-Shot models, matched EgoBlind baselines, and PAVE. Within each row, the best and second-best values per metric are marked in \textbf{bold} (darkest shade) and \underline{underlined} (lighter shade), respectively.}
\label{tab:app_question_form_full}
\small
\setlength{\tabcolsep}{6pt}
\setlength{\aboverulesep}{0pt}
\setlength{\belowrulesep}{0pt}
\renewcommand{\arraystretch}{0.95}
\begin{tabularx}{0.92\textwidth}{l *{6}{Y}}
\toprule
\multirow{2}{*}{\textbf{Question Form}}
& \multicolumn{2}{c}{\textbf{Zero-Shot}}
& \multicolumn{2}{c}{\textbf{EgoBlind}}
& \multicolumn{2}{c}{\textbf{PAVE (Ours)}} \\
\cmidrule(lr){2-3}\cmidrule(lr){4-5}\cmidrule(lr){6-7}
& \textbf{Acc. (\%)} & \textbf{BLV Score}
& \textbf{Acc. (\%)} & \textbf{BLV Score}
& \textbf{Acc. (\%)} & \textbf{BLV Score} \\
\midrule
\multicolumn{7}{@{}l}{\textit{InternVL2.5-8B}} \\
What & 44.93 & 2.5154 & \gTwo \underline{49.34} & \gTwo \underline{2.5242} & \gFive \textbf{49.78} & \gFive \textbf{2.5991} \\
Where & 38.46 & 2.2885 & \gFive \textbf{50.00} & \gTwo \underline{2.5385} & \gFive \textbf{50.00} & \gFive \textbf{2.6538} \\
How & \gTwo \underline{60.17} & \gTwo \underline{2.9068} & 52.54 & 2.6356 & \gFive \textbf{63.56} & \gFive \textbf{2.9831} \\
Which & 56.86 & 3.1373 & \gTwo \underline{64.71} & \gTwo \underline{3.2353} & \gFive \textbf{66.67} & \gFive \textbf{3.4118} \\
Other & \gTwo \underline{68.00} & \gTwo \underline{3.2400} & 64.00 & 3.2000 & \gFive \textbf{72.00} & \gFive \textbf{3.3600} \\
Yes/No & 59.01 & 3.1431 & \gTwo \underline{72.61} & \gTwo \underline{3.4488} & \gFive \textbf{77.56} & \gFive \textbf{3.7297} \\
\textbf{Overall} & 54.35 & 2.9074 & \gTwo \underline{62.88} & \gTwo \underline{3.0660} & \gFive \textbf{67.00} & \gFive \textbf{3.2878} \\
\midrule
\multicolumn{7}{@{}l}{\textit{LLaVA-OV-7B}} \\
What & 39.21 & \gFive \textbf{2.4714} & \gTwo \underline{41.85} & 2.3172 & \gFive \textbf{45.81} & \gTwo \underline{2.3612} \\
Where & 34.62 & 2.1538 & \gTwo \underline{45.19} & \gTwo \underline{2.5962} & \gFive \textbf{52.88} & \gFive \textbf{2.6442} \\
How & 46.61 & 2.5763 & \gTwo \underline{57.63} & \gTwo \underline{2.7797} & \gFive \textbf{66.10} & \gFive \textbf{3.0339} \\
Which & 52.94 & 2.8824 & \gTwo \underline{54.90} & \gTwo \underline{2.9804} & \gFive \textbf{72.55} & \gFive \textbf{3.4118} \\
Other & \gTwo \underline{60.00} & \gFive \textbf{3.0000} & 52.00 & 2.8000 & \gFive \textbf{64.00} & \gFive \textbf{3.0000} \\
Yes/No & 59.01 & 3.1820 & \gTwo \underline{70.32} & \gTwo \underline{3.4329} & \gFive \textbf{71.73} & \gFive \textbf{3.5106} \\
\textbf{Overall} & 50.96 & 2.8524 & \gTwo \underline{59.49} & \gTwo \underline{3.0147} & \gFive \textbf{63.79} & \gFive \textbf{3.1210} \\
\midrule
\multicolumn{7}{@{}l}{\textit{Qwen2.5-VL-7B}} \\
What & 43.17 & \gTwo \underline{2.4537} & \gFive \textbf{46.26} & \gFive \textbf{2.5110} & \gTwo \underline{44.93} & 2.4449 \\
Where & 28.85 & \gTwo \underline{1.9808} & \gFive \textbf{41.35} & \gFive \textbf{2.1923} & \gTwo \underline{36.54} & \gTwo \underline{1.9808} \\
How & \gFive \textbf{58.47} & \gFive \textbf{2.7627} & \gTwo \underline{50.00} & \gTwo \underline{2.5593} & 45.76 & 2.1017 \\
Which & \gTwo \underline{49.02} & \gTwo \underline{2.5490} & 47.06 & 2.5098 & \gFive \textbf{54.90} & \gFive \textbf{2.8431} \\
Other & \gTwo \underline{60.00} & \gTwo \underline{3.1200} & 52.00 & 2.7600 & \gFive \textbf{68.00} & \gFive \textbf{3.2000} \\
Yes/No & 56.18 & 2.9629 & \gTwo \underline{62.54} & \gTwo \underline{3.2191} & \gFive \textbf{65.02} & \gFive \textbf{3.3092} \\
\textbf{Overall} & 50.87 & 2.7259 & \gTwo \underline{54.81} & \gFive \textbf{2.8588} & \gFive \textbf{55.64} & \gTwo \underline{2.8478} \\
\bottomrule
\end{tabularx}
\Description{Complete question-form breakdown of Accuracy and BLV Score across InternVL2.5-8B, LLaVA-OV-7B, and Qwen2.5-VL-7B for What, Where, How, Which, Other, and Yes/No questions.}
\end{table*}

\begin{figure*}[!t]
\centering
\includegraphics[width=\textwidth]{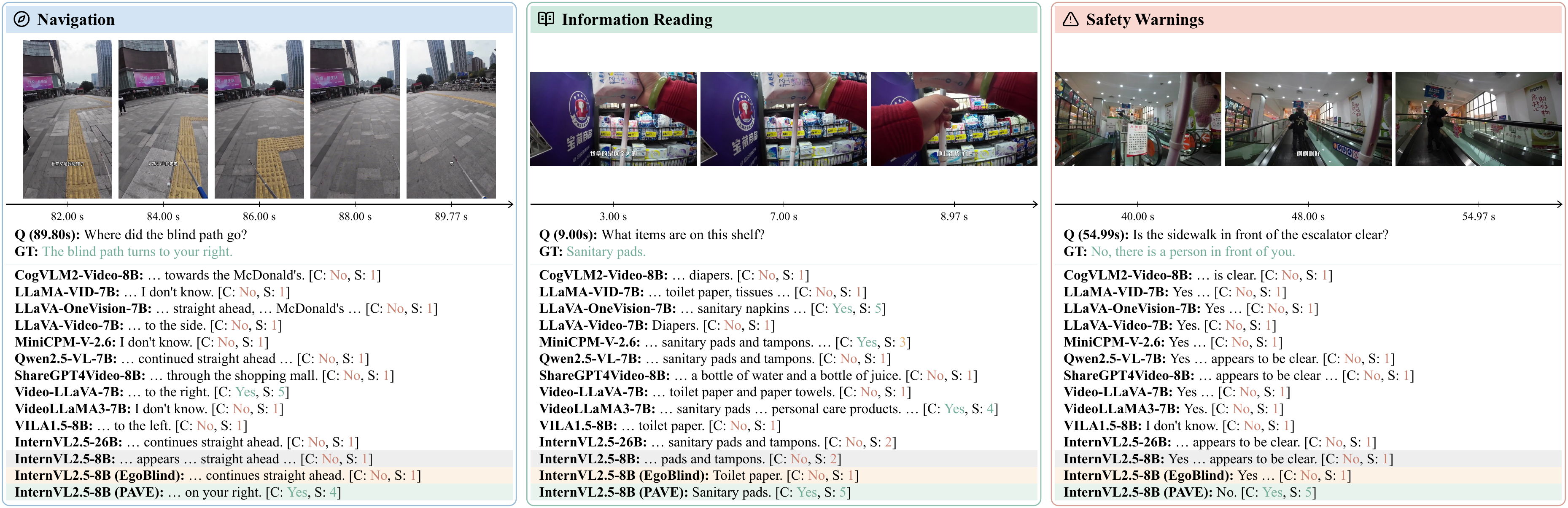}
\caption{Qualitative comparison across twelve Zero-Shot VLMs, the matched EgoBlind baseline, and PAVE on three representative egocentric queries.}
\label{fig:response_comparison}
\Description{Three video question-answer cases showing time-stamped video frames, references, model-generated answers, and correctness and assistive-quality judgments across Navigation, Information Reading, and Safety Warnings.}
\end{figure*}

\subsection{Additional Qualitative Response Comparisons}
\label{app:qualitative_responses}

Complementing the situated \emph{Tool Use} analysis in Section~5.4, Figure~\ref{fig:response_comparison} compares all twelve Zero-Shot VLMs, the matched EgoBlind baseline, and PAVE across three additional representative egocentric queries, illustrating how our three routed policies and correctness-first reward ($R_{qi} = 2C_{qi} + S_{qi}/5$) operate in practice. First, in the \textbf{Navigation} scenario (Case~1, routed to $P_{\mathrm{spatial}}$), when queried about the trajectory of tactile paving, multiple Zero-Shot VLMs and the EgoBlind baseline erroneously instruct the user to proceed straight ahead ($C_q=0$), whereas PAVE tracks the spatial progression across frames and accurately specifies that the path turns to the right of the user ($C_q=1, S_q=4$). Second, in the \textbf{Information Reading} scenario (Case~2, routed to $P_{\mathrm{evidence}}$), when identifying merchandise on a retail shelf, baseline models hallucinate unrelated items such as diapers, toilet paper, or bottled beverages ($C_q=0$), whereas PAVE grounds its readout strictly in the visible packaging text to report \emph{``Sanitary pads''} ($C_q=1, S_q=5$). Third, in the \textbf{Safety Warnings} scenario (Case~3, routed to $P_{\mathrm{decision}}$), when verifying whether the approach to an escalator is unobstructed, nearly all baseline configurations falsely report a clear path ($C_q=0, S_q\le 1$) and fail to detect the pedestrian standing directly ahead; in contrast, PAVE accurately judges the scene as unsafe and outputs a decisive, concise \emph{``No.''} ($C_q = 1, S_q = 5$), effectively prioritizing immediate user safety without unnecessary verbosity.

\section{Implementation Details}
\label{app:implementation}

\textbf{Hardware and optimization hyperparameters.} All training and evaluation experiments are conducted on compute nodes equipped with 2$\times$ NVIDIA A800 (80GB) GPUs. Both the Structured SFT and \filter{}  stages are optimized using AdamW with cosine learning rate schedules. Structured SFT uses linear warmup over the first 3\% of training steps, whereas \filter{} uses a 10-step linear warmup. During Structured SFT, we use peak learning rates of $4\times 10^{-5}$, $1\times 10^{-5}$, and $2\times 10^{-7}$ for InternVL2.5-8B, LLaVA-OV-7B, and Qwen2.5-VL-7B, respectively. During \filter{} post-training, we freeze the vision encoder and reference SFT weights ($\pi_{\mathrm{ref}}$), updating only newly attached rank-16 LoRA adapters ($\alpha=16$, dropout $0$) on the language linear projections and multimodal connector with at peak learning rate of $1\times 10^{-6}$ to preserve learned spatial-procedural priors. Online rollout generation samples $G=8$ candidate completions per query (temperature $0.8$, top-$p=0.95$) within a maximum generation length of 128 tokens, accumulating two retained contrast groups ($K(q)=1$, $|\mathcal{B}|=16$ completions) per parameter update.

\textbf{Deterministic evaluation protocol.} The reference-grounded automated evaluator used to compute binary semantic correctness $C_q\in\{0,1\}$ and holistic assistive quality $S_q\in\{0,\ldots,5\}$ during both online reinforcement learning rollouts (Section~4.3) and final benchmark evaluation is powered by gpt-4o-mini.

\bibliographystyle{ACM-Reference-Format}
\bibliography{sample-base}
\end{document}